\documentclass[letterpaper, 10 pt, journal, twoside]{IEEEtran}

\usepackage{xcolor,colortbl}
\usepackage{amsmath,amsfonts,amssymb,mathrsfs}
\usepackage{graphicx}
\usepackage[bookmarks=true]{hyperref}
\usepackage{xcolor}
\usepackage{wrapfig}

\usepackage{array}
\usepackage[linesnumbered,ruled,vlined]{algorithm2e}

\newcolumntype{C}[1]{>{\arraybackslash}p{#1}}

\newcommand{\changed}[1]{{\color{black} #1}}

\title{Task-driven Perception and Manipulation\\for Constrained Placement of Unknown Objects}

\author{Chaitanya Mitash, Rahul Shome, Bowen Wen, Abdeslam Boularias and Kostas Bekris%
\thanks{Manuscript received: Feb, 24, 2020; Revised May, 16, 2020; Accepted June, 22, 2020.}
\thanks{This paper was recommended for publication by Editor Dr. Hong Liu upon evaluation of the Associate Editor and Reviewers' comments. 
This work was supported by the NSF, grant numbers IIS-1734492 and IIS-1723869.} 
\thanks{The authors are with the Computer Science Department of Rutgers University in Piscataway, New Jersey, 08854, USA
        {\tt\footnotesize \{cm1074, rs1123, bw344, ab1544, kb572\}@rutgers.edu}}%
\thanks{Digital Object Identifier (DOI): see top of this page.}
}

\newcommand{\object}{O}

\newcommand{\seen}{\mathcal{S}}
\newcommand{\unseen}{\mathcal{U}}
\newcommand{\gripper}{E}
\newcommand{\segment}{s}

{\end{list}}

\newcommand{\pose}{P}

\newcommand{\vol}{O^*}
\newcommand{\pinit}{\pose_{\rm init}}

\newcommand{\graspset}{\mathcal{G}}

\begin{document}

\maketitle

\begin{abstract}
Recent progress in robotic manipulation has dealt with the case
of \changed{previously unknown objects} in the context of relatively
simple tasks, such as bin-picking. Existing methods for more
constrained problems, however, such as deliberate placement in a tight
region, depend more critically on shape information to achieve safe
execution. \changed{This work deals with pick-and-constrained
placement of objects without access to geometric models. The objective
is to pick an object and place it safely inside a desired goal region
without any collisions, while minimizing the time and the sensing
operations required to complete the task. An algorithmic framework is
proposed for this purpose, which performs manipulation planning
simultaneously over a conservative and an optimistic estimate of the
object's volume. The conservative estimate ensures that the
manipulation is safe while the optimistic estimate guides the
sensor-based manipulation process when no solution can be found for
the conservative estimate. To maintain these estimates and dynamically
update them during manipulation, objects are represented by a simple
volumetric representation, which stores sets of occupied and unseen
voxels.  The effectiveness of the proposed approach is demonstrated by
developing a robotic system that picks a previously unseen object from
a table-top and places it in a constrained space. The system comprises
of a dual-arm manipulator with heterogeneous end-effectors and
leverages hand-offs as a re-grasping strategy. Real-world experiments
show that straightforward pick-sense-and-place alternatives frequently
fail to solve pick-and-constrained placement problems. The proposed
pipeline, however, achieves more than 95\% success rate and faster
execution times as evaluated over multiple physical experiments.}

\end{abstract}


\section{Introduction}
\label{sec:intro}
\IEEEPARstart{O}{bject} placement in tight spaces is a challenging problem in robot
manipulation. In contrast to a simpler pick-and-drop problem, only
specific object poses will allow it to fit in a tight space. Such
scenarios occur in logistics applications, such as packing items into
boxes, or in service robotics, such as inserting a book into a gap in
a bookshelf. Recent work has focused on variants of this problem, such
as bin-packing~\cite{wang2019robot, shome2019towards} and table-top
placement in clutter~\cite{haustein2019object}. Nevertheless, in many
cases a geometric and textured 3D model for the manipulated object is
assumed to be known. Possessing such high-fidelity models is expensive
both in terms of time and effort. In several setups it becomes
infeasible to build models due to the wide variety of objects to be
manipulated and the resources required for obtaining the models. Some
recent robot manipulation systems~\cite{mahler2019learning,
zeng2018robotic} have shown the capacity of picking novel and
previously unseen objects from clutter. These systems, however,
typically assume no constraints for the object's placement. Therefore,
the object is grasped with any feasible and stable grasp without
reasoning about placement. Some alternatives do not require exact
models of objects but operate with category-level prior
information. Examples include an approach based on sparse keypoint
representations~\cite{manuelli2019kpam} and deep reinforcement
learning~\cite{Gualtieri:2018aa}. While the employed representations
can guide manipulation planning solutions, they do not account for
safety as they do not consider geometric or physical constraints.

\begin{figure}[t]
\centering
\includegraphics[width=0.9\linewidth, keepaspectratio]{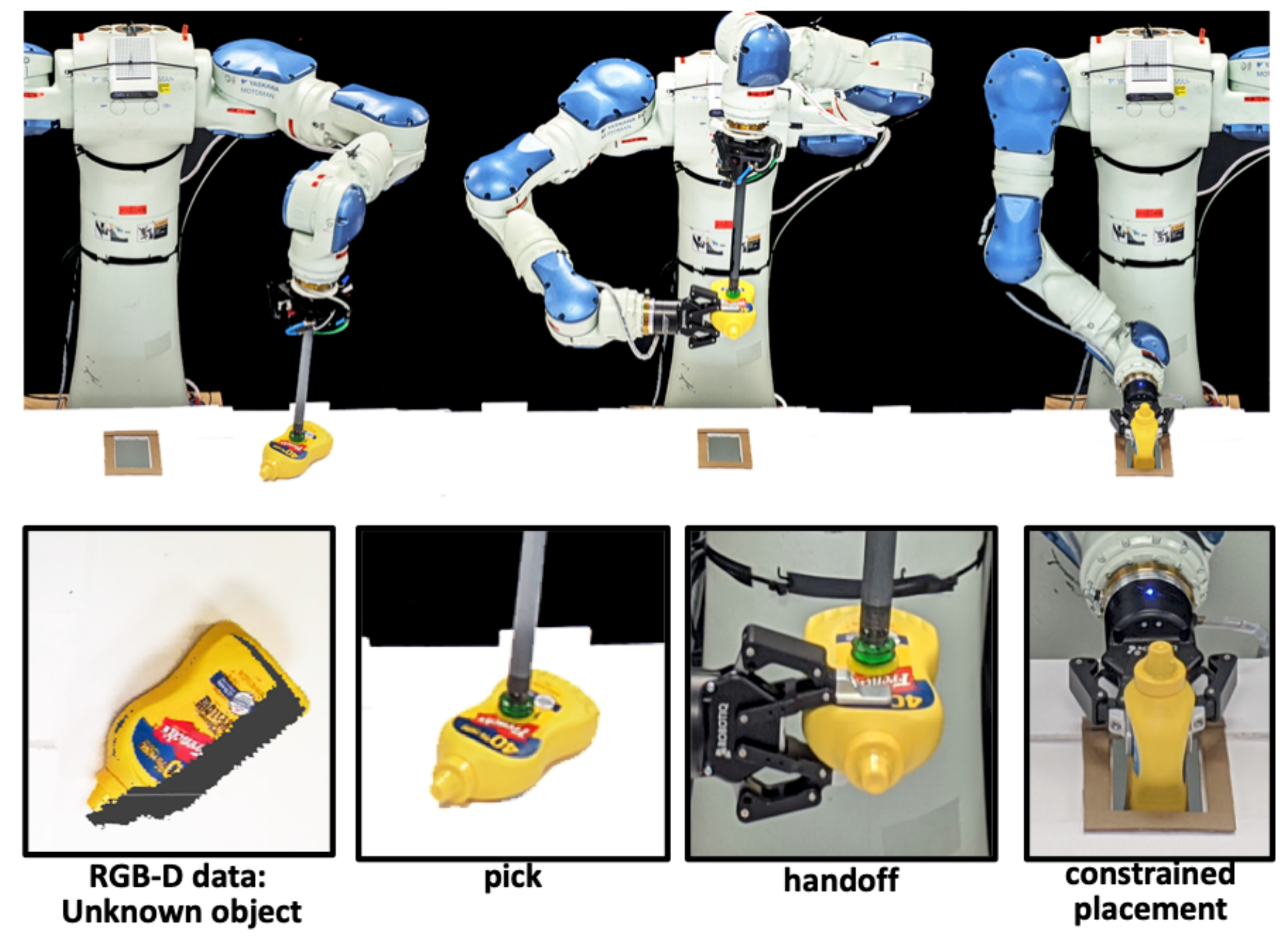}
\vspace{-.1in}
\caption{\changed{A demonstration of pick-handoff-place returned by
the proposed framework for inserting a previously unknown object in a
constrained space. This work does not focus on the last step of
precise, closed-loop insertion but how to reason about the object's
shape so as to safely bring it to the opening of the placement
area. Only a subset of placement poses allow the object to fit into
the target area. Experiments (Table.\ref{table:task_succ}) consider a larger margin than
the demonstration shown here (2 cm instead of 1 cm shown)}.}
\vspace{-.12in}
\label{fig:intro1}
\end{figure}

\changed{This work targets pick-and-place problems where
the task imposes constraints on the placement pose. The capabilities
of a manipulator impose limitations on what placement poses are
reachable depending on the grasp, making certain grasps more desirable
than others. This requires careful reasoning to select the pick that
will allow the desired placement. This will be referred to as the {\it
pick-and-constrained-placement} problem. In the context of this
problem, it is possible that a feasible placement pose is not directly
attainable using a pick-and-place operation. Instead, it may require a
re-grasping of the object or a hand-off to be executed.

Solutions to such problems typically need object models for collision
checking, which this work does not assume. Picking, placement, and
re-grasping actions need to be computed given partial viewpoints of
the object acquired from the sensor, as in Fig.~\ref{fig:intro2}. This
work approaches the pick-and-constrained-placement problem without
prior object models as {\it integrated perception and manipulation
planning}. The objective is to place the entire object safely inside
the desired goal region without any collisions, while minimizing the
time and sensing operations required to complete the task.

\begin{figure}[t]
\centering
\includegraphics[width=0.9\linewidth, keepaspectratio]{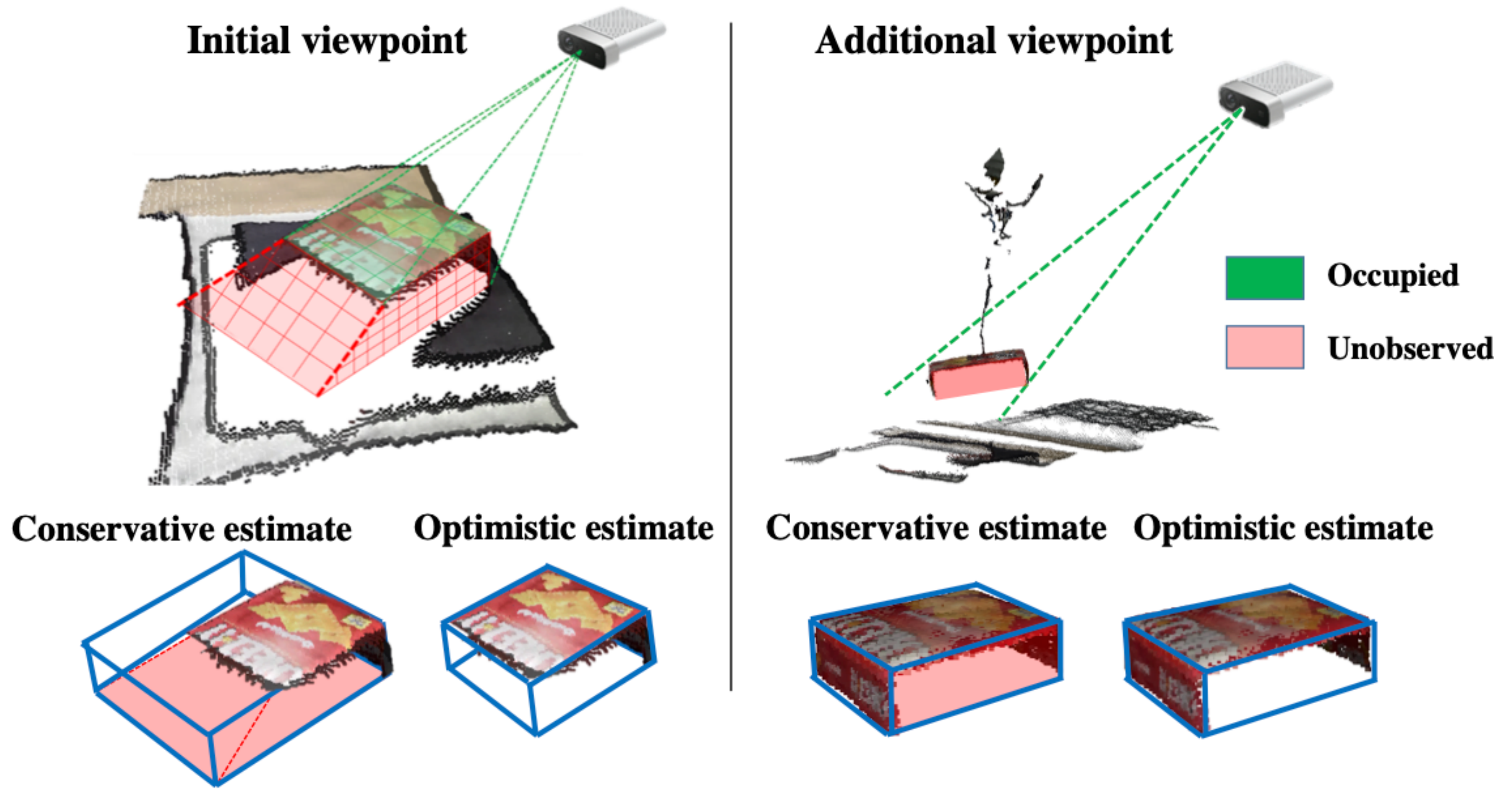}
\vspace{-.1in}
\caption{\changed{(Left) Figure shows a partial view of the object
from the sensor. A {\it conservative} estimate of the object considers
both the observed and the unobserved parts of the object given knowledge
of the support surface. An {\it optimistic} estimate considers only
the observed parts. (Right) The estimates are updated as the object is
manipulated to different sensor viewpoints.}}
\vspace{-.2in}
\label{fig:intro2}
\end{figure}

One option is to pick the object with a task-agnostic grasp that
solely interacts with the visible part of the object as in
pick-and-drop systems~\cite{mahler2019learning,
zeng2018robotic}. After picking, the object's shape can be completely
reconstructed by manipulating it to different configurations in front
of the sensor. Then, geometric planning can be performed based on the
reconstructed model. Nevertheless, not only will this option be very
time-consuming given the object must be moved to multiple viewpoints,
but using a goal-agnostic pick may not allow the constrained placement
without multiple regrasps. A key point of this work is that
constrained placement task can be successfully completed without a
complete object model. This motivates a dynamic estimate of the
object's shape to solve the problem, and a planning approach capable
of using and updating such a representation on the fly.

The algorithmic solution proposed simultaneously operates over {\it
conservative} and {\it optimistic} estimates of the object's 3D
volume, as in Fig.~\ref{fig:intro2}. The {\it conservative} estimate
considers the entire volume attached to the object, which has not been
observed by the sensor as part of the object. The {\it optimistic}
estimate considers only the object's observed region to be its
complete representation. While the {\it conservative} estimate ensures
that the manipulation is safe, the {\it optimistic} estimate guides
the action selection when no solution can be found for the {\it
conservative} estimate. Both estimates are dynamically updated by
incorporating new viewpoints, which are selected such that a
safe-to-execute constrained placement solution can be found with
minimal sensing.

To efficiently obtain these dynamic estimates, this work proposes to
utilize a simple volumetric representation. Similar to
occupancy-grids \cite{moravec1985high} often used in the context of
robot navigation to store the occupied and free space, this
representation stores whether a voxel in the object's reference frame
is occupied, unoccupied or unobserved. Instead of utilizing fixed-size
grids or octrees to store the volumetric information, the
representation maintains sets of occupied and unobserved voxels. This
minimalistic representation provides efficiency at the cost of
building exact models but proves to be sufficient to solve the
considered problem.}

The effectiveness of the proposed approach is demonstrated by
developing a robotic system that picks a previously unseen object from
a table-top and places it in a constrained space (Fig.~\ref{fig:intro1}). The system comprises of a dual-arm manipulator, an RGB-D sensor, a vacuum-based end-effector and an
adaptive, finger-based hand. Additionally, the system features {\it
handoffs} to transfer objects between the two arms and a strategy
to adjust the computed motion trajectories during real-world execution
given sensing updates. Handoff is a re-grasping strategy that allows
more flexibility in solving constrained placement
problems. Closed-loop execution handles stochastic in-hand motions of
objects resulting from unmodeled physical forces like gravity, inertia
and grasping contacts.

\changed{$240$ real-world manipulation experiments are performed to
compare the proposed solution and the straightforward
pick-sense-and-place alternative \footnote{Videos and supplementary
algorithmic description: \url{https://robotics.cs.rutgers.edu/task-driven-perception/}}.} The
experiments demonstrate that the proposed pipeline is both robust and
efficient in handling objects with no prior models within the
limitations of the end-effector and the sensor. It achieves a success
rate of 95.82\%, which is much higher than an alternative that commits
to a pick without manipulation planning and performs object
reconstruction from heuristic viewpoints without utilizing the
conservative volumetric representation. The proposed pipeline results
in fewer sensing operations and achieves faster execution times.

\section{Related Work}
\label{sec:related}
\begin{figure*}[t]
\centering
\includegraphics[width=0.95\textwidth, keepaspectratio]{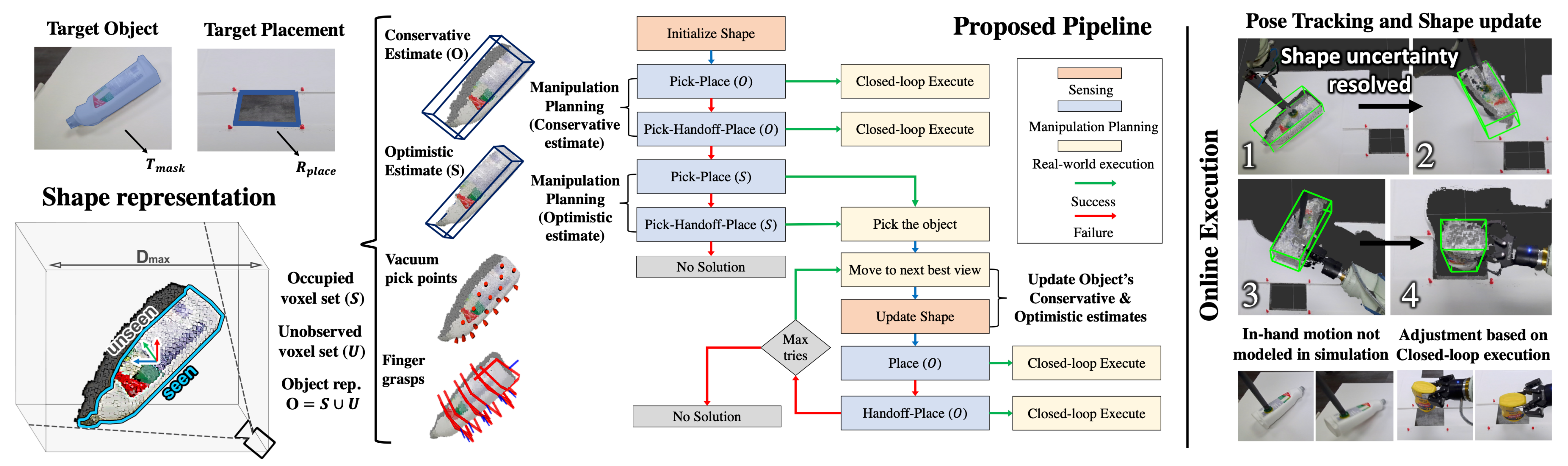}
\vspace{-0.1in}
\caption{\changed{(left) The proposed framework considers as input
    RGB-D images and the target object mask and builds a shape
    representation based on the observed and the occluded part of the
    object.  (center) It simultaneously operates over a conservative
    and an optimistic estimate of the object's volume to compute a
    sequence of manipulation and sensing actions for
    pick-and-constrained-placement. The object is dynamically updated
    during the manipulation until a safe to execute sequence of action
    is available. (right) Online adaption is performed, which is
    informed by pose-tracking to counter the effect of stochastic
    in-hand motion of the object, which is not modeled during
    planning.}}
\vspace{-0.1in}
\label{fig:pipeline}
\end{figure*}

\changed{This section discusses  existing pick-and-place
manipulation pipelines, object representations for these pipelines and
assumptions about the object's shape and category.

\noindent{\bf Manipulation pipelines for pick-and-place:}} Given access
to object models, previous work has addressed problems such as
bin-picking \cite{Princeton}, tight-packing \cite{wang2019robot,
shome2019towards} and placement of grasped objects in
clutter \cite{haustein2019object}. Most manipulation pipelines for
novel objects \cite{mahler2019learning, zeng2018robotic} focus on
picking the object but do not address the problem of constrained
placement. It has been demonstrated that robust grasps can be
computed \cite{Gualtieri:2017aa} over 3d point cloud representations
of novel objects by learning local geometric features. However,
constrained placement tasks require simultaneously evaluating
placements and grasps over the objects, which is a relatively harder
problem than task-agnostic grasping. A recent
work, \cite{Gualtieri:2018aa} performs pick-and-place of objects
without object models, but within a single category, by training an
end-to-end deep reinforcement learning framework within the task
context. Given that it is hard to interpret the learned policies, it
is not clear how the policies learned with rewards coming from a
specific task can be generalized to other similar tasks,
configurations and objects. Another recent
effort \cite{manuelli2019kpam} proposes using semantic keypoints as
category-level object representation in conjunction with shape
completion \cite{gao2019kpam} to model collision
geometry. Nevertheless, such techniques typically require access to
prior knowledge of the object's category to complete its shape and the
output is often too noisy for safe manipulation planning in
constrained spaces. This work does not assume any knowledge of the
object's geometry or category prior while solving pick-and-constrained
placement problems.

\changed{\noindent{\bf Object representation for manipulation:} Objects
are often represented as mesh models that capture the surface of the
object. The models are built either using a turntable
setup \cite{calli2015yale}, or via in-hand scanning by a human
user \cite{wang2019hand} or a robotic
arm \cite{krainin2010manipulator}. A popular technique for surface
reconstruction is Truncated Signed Distance Function
(TSDF) \cite{curless1996volumetric, newcombe2011kinectfusion} which
fuses multiple depth observations from a sensor and maintains a signed
distance to the closest zero-crossing (representing the
surface). Alternatively, the Surfel
representation \cite{weise2009hand} is used to store local surface
patches with position and normal information. Nonetheless, the
objective is to generate complete meshes and often involves additional
setup and post-processing steps.} The complete models are then used to
perform pose estimation \cite{xiang2017posecnn, mitash2018robust, wen2020robust, mitash2020scene} over
the online sensor data and transfer the manipulation actions that are
defined over the model to the scene. Given the effort in modeling
every object instance, some approaches operate at the category-level
where objects are represented in a normalized object
frame \cite{wang2019normalized} or via a canonical
model \cite{rodriguez2018transferring}. But given large intra-class
shape variation in certain scenarios, it is hard to capture the shape
in a single category-level pose representation. This often leads to
planning manipulation actions that end up in physically-unrealistic
configuration for certain instances of the category.

An alternative is volumetric shape completion that has been
studied in the context of grasping \cite{varley2017shape,
bohg2011mind, quispe2015exploiting}, manipulation \cite{gao2019kpam}
and object search \cite{price2019inferring}. \changed{These approaches
come up with a most-likely estimate of the object from a partial view
based on assumptions, such as symmetry or category-level
information. Operating over such estimates can lead to collisions if
the estimated volume is smaller than the actual object. Instead, the
proposed approach operates only over the sensor data without any
assumptions about the object's shape. The object representation in
this work is most similar to occupancy grids. Occupancy grids are
often used in the context of SLAM or indoor navigation to map
boolean or probabilistic occupancy properties either over fixed grid
structures \cite{moravec1985high} or over a more efficient octree
representation \cite{hornung2013octomap}. Instead of a fixed grid
structure the current representation stores the {\it
occupied} and {\it unobserved} voxels of the object as sets. These sets are
updated based on new viewpoints. The minimalistic representation can
efficiently maintain a dynamic representation of the conservative and
optimistic object volume. Thus, the representation is utilized in the
context of the pipeline to perform manipulation planning directly over
sensor data, without any assumptions over geometric or category-level
priors.}

\section{Problem Setup and Notation}
\label{sec:problem}
\changed{This section formulates the {\it integrated perception and manipulation planning problem} for constrained placement.}

\noindent {\bf Object representation:} A rigid object can be defined by a
region occupied by the object $\vol \subset \mathbb{R}^3$ in its local
reference frame that represents its shape. Given a pose $\pose \in
SE(3)$, the region occupied by the object at $\pose$ is denoted by
$\vol_{\pose}$. \changed{It should be noted that a geometric model is
not available for the object to be manipulated}, i.e., $\vol$ is
unknown. Thus, $\object$ defines an object representation over which
manipulation planning can operate. In general,
$\object \neq \vol$. $\object$ is derived from an initial view of the
object \changed{given point cloud and image segmentation. The
resulting object model is typically incomplete, and may not be
sufficient to safely place the object in a constrained area.}

\noindent {\bf Constrained placement:} Given an object at an initial
pose $\pinit\in SE(3)$, the goal of the constrained placement problem
is to transfer $\vol$ to a pose $\pose_{\rm target} \in
SE(3)$, such that $\vol_{\rm \pose_{\rm target}} \subset R_{\rm
place}$ where $R_{\rm place} \subset \mathbb{R}^3$ is the target
placement region.

\noindent \textbf{Manipulation Planning: }
Manipulation planning for constrained placement involves computing a sequence of manipulation actions (picks, placements, re-grasps) that can move the object $\vol$ from $\pinit$ to $\pose_{\rm target}$, which successfully solves a constrained placement task. Such a solution consists of motions of the arms denoted by $\Pi$ parameterized by the time of the motions. $\Pi(0)$ is the initial arm configuration, and $\Pi(1)$ has an arm placing the object at $\pose_{\rm target}$.

\changed{\noindent \textbf{Integrated Perception and Manipulation Planning: }
Given that the true object geometry $\vol$ is unknown and planning can
make use only of the partial object representation $\object$, {\it
perception actions} are also necessary. These actions can update the
object representation $\object$ by manipulating it to desirable
configurations in front of the sensor and obtaining additional sensing
information. Thus, the problem involves computing a sequence of
perception and manipulation actions, such that: i) the object after
executing the sequence of actions ends up inside the defined
constraints, i.e., $\vol_{\pose_{\rm final}}$ is within $R_{\rm
place}$, where $\pose_{\rm final}$ is the resultant pose of the object
after applying the actions; ii) and the returned sequence of
perception and manipulation actions minimizes the task execution
time.}

\section{Proposed pipeline}
\label{sec:approach}

This section presents the proposed pipeline as shown in
Figure.~\ref{fig:pipeline}. Given as input RGB-D image of the scene
and the target object mask $T_{\rm mask}$, the object representation
$\object$ is initialized with it's origin at the centroid of the 3D
point cloud segment corresponding to $T_{\rm mask}$ and the reference
frame at identity rotation with respect to the camera frame. Within a
voxel grid centered at the origin, each voxel is labeled as either 1)
\textit{observed and occupied} $\seen$, 2) \textit{unobserved}
$\unseen$, or 3) observed and unoccupied, i.e., empty voxels that are
implicitly modeled as a set of voxels $\{p \in \mathbb{R}^3 \mid p
\notin \seen \cup \unseen, \|p - \textrm{origin}(\object)\| < D_{\rm
  max} \}$, for a maximum dimension parameter $ D_{\rm max} = 30
cm$. The representation is stored as a set $\object$ that
consists of two mutually exclusive sets of voxels $\seen$ and
$\unseen$ in $\mathbb{R}^3$. $\seen$ is a set of \textit{occupied} voxels
on the surface of the object that are observed by the RGB-D
sensor. $\unseen$ is a set of \textit{unobserved} voxels in space that
have not been observed by the sensor given the viewpoints but
\textit{have a non-zero probability of belonging to the target
  object}. Thus $\object = \seen \cup \unseen$, where,
$\seen~\cap~\unseen = \phi$.

\changed{

A set of grasps and placements are computed simultaneously over a {\it
  conservative estimate} and an {\it optimistic estimate} of the
object's volume. The {\it conservative estimate} corresponds to
$\object$, while the {\it optimistic estimate} only considers the
observed part of the object, i.e., $\seen$. Manipulation planning is
performed considering the grasps and placements computed over
$\object$. The objective is to compute a sequence of these
manipulation actions and corresponding arm motions, which allow to
connect a grasp to a placement pose. Any manipulation planning
solution computed over $\object$ can be directly executed in the
real-world as it is necessarily collision-free with respect to the
true object shape $\vol$, given that $\vol \subseteq \object$. Often
no solution can be found for the task as $\object$ may significantly
overestimate $\vol$. In such a scenario, the object is picked and
manipulated to acquire new observations, thereby updating $\object$.

The choice of picking point is critical as it might influence the
solution once $\object$ has been updated. For this reason,
manipulation planning is performed over the optimistic estimate of the
object's volume. In this case, all actions after picking, such as
re-grasps and placements are computed over $\seen$. If no placements
are achievable given $\seen$, the problem is \textit{not solvable},
since $\seen \subset \vol$. If a solution is found for $\seen$, it
informs the selection of the picking point over $\object$.

The next decision is the selection of the next best view. It is
selected among a set of pre-defined discrete viewpoints with an
objective of exposing the highest number of unobserved voxels in
$\unseen$. This is found by rendering $\seen$ at each of the
viewpoints and computing the number of voxels in $\unseen$ that are
visible, given the rendered image. The selected viewpoint is
most-likely to reduce the conservative volume of the object. The
object is then moved to this viewpoint and $\object$ is updated.

The size of the set $\object$ (and thus the conservative volume) is
largest at initialization. Any update to $\object$ either removes a
point $p \in \unseen$ (if it is observed to be empty) or $p$ can be
moved from $\unseen$ to $ \seen$. To update $\object$, the observed
segment $\segment^{t}$ at time $t$ is transformed to the object's
local frame based on the estimated pose $\pose ^ {t}$. For each point
$p$ on the transformed point cloud, its nearest neighbor $p^\seen \in
\seen$ and $p^\unseen \in \unseen$ are found. If $\mid p^\seen - p
\mid < \delta_c$ where $\delta_c$ is the correspondence threshold, $p$
is considered to be already present. Otherwise, if $\mid p^\unseen - p
\mid < \delta_c$, $p^\unseen$ is removed from $\unseen$ and added to
$\seen$. Finally, the method iterates over all points in
$\unseen_{\pose_t}$ to remove points in $\unseen$, which belong to the
empty part of space based on the currently observed depth
image. Applying these constraints in the update significantly reduces
the drift that occurs in simultaneous updates to the object's pose and
shape.

Grasps and placements are re-computed over the updated object estimate
and manipulation planning is performed again. This process is repeated
until either a solution is found for the constrained placement task or
the algorithm runs out of a maximum number of trials. This means that
the pipeline does not require the object to be completely
reconstructed, but only enough to compute a safe-to-execute solution
for the placement task.

\section{System Design \& Implementation}}

\begin{wrapfigure}{r}{0.24\textwidth}
\vspace{-0.2in}
\centering
\includegraphics[width=0.24\textwidth, keepaspectratio]{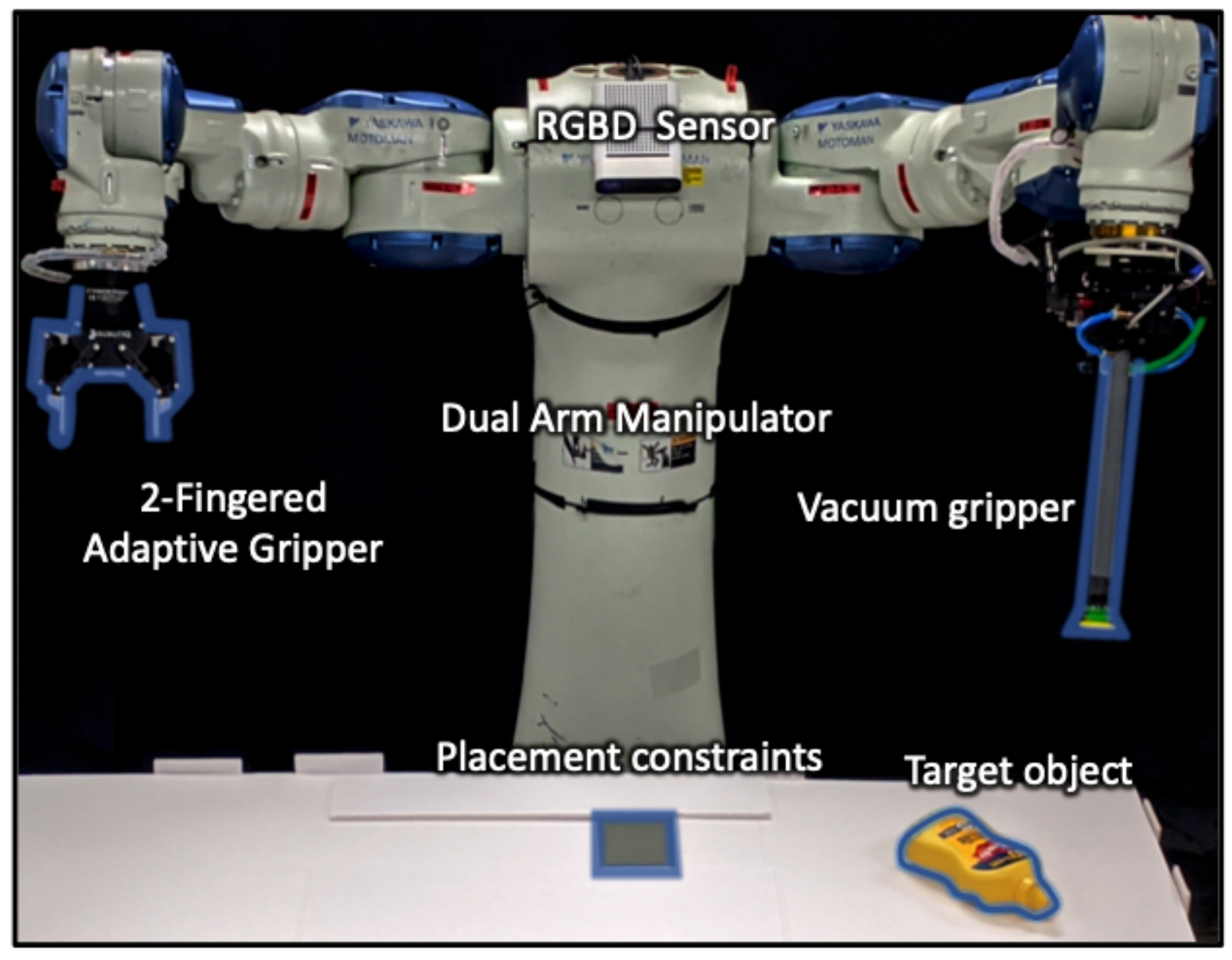}
\vspace{-0.25in}
\caption{\changed{Hardware Setup.}}
\vspace{-0.1in}
\label{fig:system}
\end{wrapfigure}

Fig.~\ref{fig:system} shows the hardware setup. It comprises a
dual-arm manipulator (Yaskawa Motoman) with two 7-dof arms. The left
arm is fitted with a narrow, cylindrical end-effector with a vacuum
gripper; and the right arm is fitted with a Robotiq 2-fingered
gripper. A single RGB-D sensor (Kinect Azure) is mounted on the robot
overlooking both the picking and the placement regions. The sensor is
configured in Wide-FOV mode to capture images at 720p resolution with
a frequency of up to 20Hz. Below are the implementation details
corresponding to different components of the proposed pipeline for
this hardware setup.

{\bf Grasp computation: } Grasp sets $\graspset_{\rm l}$ and
$\graspset_{\rm r}$ are computed over the object representation
$\object$ by ensuring stable geometric interaction with the observed
part of the object $\seen$ and being collision-free with \textit{both
$\seen$ and $\unseen$}, thereby ensuring \textit{safe} and successful
execution. It is also crucial for the success of manipulation planning
to have large, diverse grasp sets at its disposal. This is distinct
from the typical objective of grasp generation modules that primarily
focus on the quality of the top (few) returned grasps. For instance,
in Fig.~\ref{fig:pipeline}, the grasps are spread out over
$\object$ with different approach directions, which provide options to
manipulation planning and aid solution discovery.

Vacuum grasps $\graspset_{\rm l}$ are computed by uniformly sampling
pick points and their surface normals from $\seen$, and ranked in quality by
their distance from the shape centroid. The grasp set $\graspset_{\rm
r}$ for the fingered gripper samples a large set of grasps over
$\object$ according to prior work \cite{Gualtieri:2017aa}. Sampled
grasps are pushed forward along the grasp approach direction
until the fingers collide with points from $\seen$ or
$\unseen$, and ranked by the alignment between the finger and contact
region on $\seen$.

{\bf Placement Computation:} Given the placement region $R_{\rm
place}$, and the object representation $\object$, two boxes are
computed, 1) the maximum volume box $B_{\rm place}$ within $R_{\rm
place}$ and 2) the minimal volume box $B_{\rm \object}$ that encloses
$\object$. Candidate placement poses correspond to configurations of $B_{\rm \object}$ which fit within
$B_{\rm place}$. A discrete set of configurations ($= 24$) for the box is
computed by placing $B_{\rm \object}$ at the center of $B_{\rm place}$
and validating all axis-aligned rotations. Any pose in the returned set $\mathcal{P}_{\rm place}$ is a candidate $\pose_{\rm target}$.

{\bf Manipulation Planning: } The input to manipulation planning is
the estimated object representation $\object$, the grasp sets for
both arms, $\graspset_{\rm l}, \graspset_{\rm r}$, and the placement
poses $\mathcal{P}_{\rm place}$. Manipulation planning returns a
sequence of prehensile manipulation actions that ensure a collision
free movement ($\Pi$) of the arms and $\object$ such that the object
is transferred from $\pinit$ to some $\pose_{\rm
target} \in \mathcal{P}_{\rm place}$. In the absence of any errors,
the execution of these actions solves the constrained placement task.

As a part of the task planning framework, a probabilistic
roadmap~\cite{Kavraki1996Probabilistic-R} consisting of $5000$ nodes
is constructed using the $\mathtt{PRM}^*$
algorithm~\cite{Karaman2011Sampling-based-} for each of the arms. The
grasps and placements for each arm can be attained by corresponding
grasping, and placement configurations of the arms, obtained
using \textit{Inverse Kinematics} solvers. Beginning with the initial
configuration of the arms, the high-level task planning problem
becomes a search over a sequence of the manipulation actions,
achievable by the \textit{pick, place or handoff} configurations. This
is described in the form of a forward search
tree~\cite{Hauser2011Randomized-Multi-Modal-} which operates over the
same roadmap~\cite{vega2016asymptotically} by invalidating
edges (motions) that collide with the object, or the other arm. The
search tree is further focused by only expanding \textit{pick-place}
and \textit{pick-handoff-place} action sequences. Each such sequence
can be achieved through a combination of different choices
of \textit{grasping, handoff, and placement} configurations. The
search traverses the set of options for grasps in the descending order
or quality, and returns the first discovered solution that
successfully achieves a valid target placement ($\pose_{\rm
target} \in \mathcal{P}_{\rm place}$).

{\bf Shape and Pose Tracking: } The object pose $\pose^t$ changes over
time with the gripper manipulating it, where $\gripper^t \in SE(3)$
denotes the gripper pose at time $t$. Between consecutive timestamps
for a perfect prehensile manipulation, $\Delta \pose^{t-1:t} = \Delta
E^{t-1:t}$ which is the change in the gripper's pose. Tracking is
introduced to account for non-prehensile within-hand motions which
violates this nicety.

The object segment at any time $\segment^t$ is computed from a) points
lying in a pre-defined region of interest in the
reference frame of the gripper, and b) by eliminating the points
corresponding to the gripper's known model. Object pose update $\Delta \pose^{t-1:t}$ is computed in three
steps:

1) Assuming rigid attachment of the object with the end-effector, the
transformation, $\Delta \gripper^{t-1:t}$ is applied to the object
segment in previous frame $\segment^{t-1}$ to obtain the expected
object segment at time t, $\segment'^{t}$.

2) To account for any within hand motion of the object, a
transformation is computed between $\segment'^{t}$ and the observation
$\segment^{t}$ via ICP. While $\Delta \pose^{t-1:t}
= \Delta \gripper^{t-1:t} * \Delta \pose_{\rm ICP}$ provides a good
estimate of relative pose between consecutive frames, accumulating
such transforms over time can cause drift.

3) A final point-set registration process is utilized to locally refine
the pose. An ICP registration step with a strict correspondence
threshold is performed between the object representation ($\object$)
at pose $\pose^{t} = \pose^{t-1} \cdot \Delta \pose^{t-1:t}$, and the
current observation $\segment^{t}$. The resulting transformation is
applied to $\Delta \pose^{t-1:t}$, and correspondingly $\pose^t$.

During manipulation, when a new viewpoint is encountered, the output
of pose tracking is utilized to update the object's shape which
assists tracking in future frames.

\changed{{\bf Reaction to Sensing Updates: } Given a manipulation
  planning solution $\Pi$, the objective is to ensure that any errors
  in execution or non-prehensile grasping interactions are
  addressed. At any point in time $t$, $\Pi(t)$ describes how the arms
  are configured. Assuming prehensile grasps, the expected object pose
  $\pose^{t*}$ can be estimated. Tracking returns the current estimate
  $\pose^{t}$. If $\pose^t \neq \pose^{t*}$ the remainder of the
  motion has to be adjusted to account for $\Delta\pose = \pose^{t*} -
  \pose^t $. Large $\Delta\pose$ errors may require complete
  re-planning of $\Pi$. In this work these adjustments are performed
  before \textit{handoffs}, and \textit{placements} by locally
  adapting $\Pi$.}

\begin{figure}[h]
\centering
\includegraphics[width=0.9\linewidth, keepaspectratio]{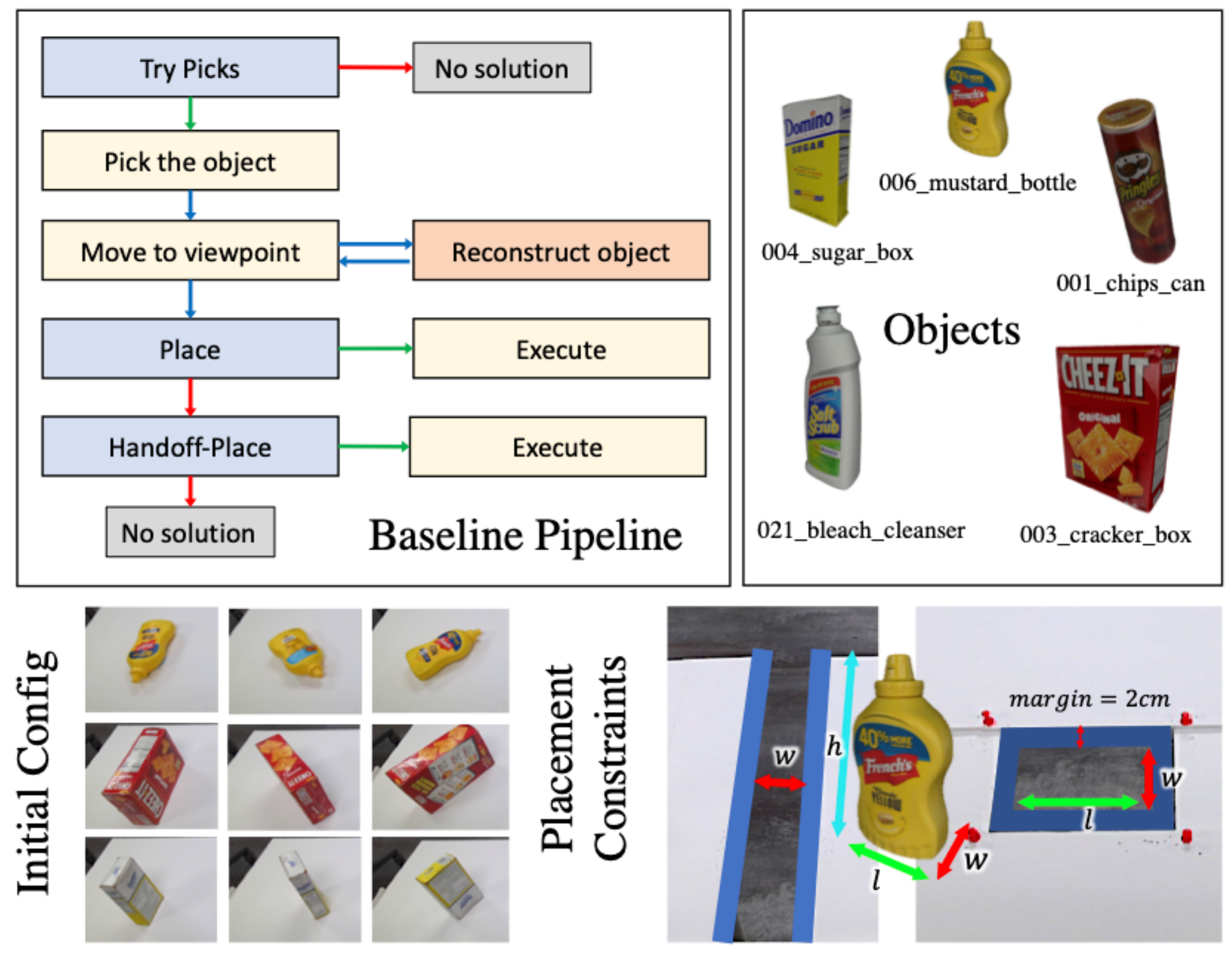}
\vspace{-0.1in}
\caption{\changed{Objects used in the experiments (top). Examples of initial configurations (left). Examples of placement constraints (right).}}
\vspace{-0.2in}
\label{fig:exp_setup}
\end{figure}

\changed{\section{Experimental Setup}}
\label{sec:system}
This section describes the setup for the experiments performed to
measure the efficacy of the proposed pipeline and the developed system
in solving the {\it pick-and-constrained-placement} problem. Given the
dual-arm manipulator, objects are placed on a table-top in front of
the left arm (vacuum gripper), with the target placement region
centrally aligned in front of the robot, reachable by both arms. The
constrained placement solutions can therefore involve a direct
placement by the left arm, or a handoff-placement with the right
arm. Following describe the different parameters of the setup followed
by the evaluation metrics.

\textit{Objects:} Experiments are performed over 5
YCB~\cite{calli2015yale} objects (Fig.~\ref{fig:exp_setup}) of
different shapes and sizes. It should be noted that \textit{no models}
are made available to the method.

\textit{Initial Configuration: } For each stable resting pose of the
object in front of the left arm, rotations were uniformly sampled
along the axis perpendicular to the plane of the table. Different
initial configurations of the object will affect the nature of the
task planning solution by virtue of a) different available initial
picks, and b) different conservative shape representation based on how
much of the object is unseen at the configuration. Configurations with
limited reachable grasps are ignored. \changed{The height of the table
is known in advance and it is used to obtain the initial point cloud
segment for the object.}

\textit{Placement Region: } An opening is created on the table surface
where the object needs to be \textit{placed}. This corresponds to the
placement task. Two placement scenarios are evaluated as shown in
Fig.~\ref{fig:exp_setup}~(bottom right). Using the measures of three
canonical dimensions measured from the object, the first class of
opening size allows four out of six approach directions for placement
to fit, while the other only allows two approach directions. An error
tolerance of $2.00 cm$ is considered in the dimension of the
opening. The idea is that more constraints (lesser approach
directions) need deliberate planning to choose precise grasp and
handoff sequences that allow the placement. \changed{Evaluating the
insertion with a lower margin would need to consider the sensor
accuracy, errors in detection of the target placement region and the
accuracy of an insertion controller, which are not the focus of this
work.}

\noindent{\bf Evaluation metrics: } \changed{Given the task, the
pipeline is responsible to pick the object and re-configure it such
that it ends up within the desired placement region on top of the
table. A simple control strategy is used to insert the object into the
hole and measure the success of the task. It utilizes cartesian
control to incrementally lower the object until the joint-limits are
reached or a collision is observed. The object is then dropped.} {\it
Success (S)} denotes the percentage of trials that result in
collision-free, successful insertion of objects within the constrained
opening, while {\it Marginal Success (MS)} records trials where the
object grazes the boundaries of the constrained space during a
successful insertion. In terms of quality metrics, {\it Task planning
time} records open-loop manipulation planning, {\it Move time} records
the time the robot is in motion, and {\it Sensing actions} counts the
number of times the robot actively re-configures the object to acquire
sensor data from a new viewpoint.

\changed{\noindent{\bf Baseline - Complete Shape Reconstruction: }} The
baseline (shown in Fig.~\ref{fig:exp_setup}) picks the object with a
{task-agnostic pick} (i.e., any pick that works) and reconstructs the
entire object by moving to pre-defined viewpoints. Manipulation
planning is performed on the reconstructed shape to find and execute a
solution for constrained placement.

A drawback of this approach is that \textit{committing to a
task-agnostic pick} might preclude solutions, which might have been
possible with a different pick. For instance, the initial pick might
not allow a direct placement or in some cases even obstruct
handoffs. Another drawback is that the amount of object reconstruction
required depends on the task. It can be inefficient to fully
reconstruct the object if a \textit{robust solution with partial
information} can be found. Finally, even with a large number of
perception actions, some parts of the objects might be missing, which
can still lead to execution failures. For instance, this can happen if
say the bottom surface is not reconstructed and fingered grasps
interact with the unmodeled part of the object during execution.

\changed{\section{Results}}
\label{sec:evaluation}
\begin{figure}[t]
\centering
\includegraphics[width=0.98\linewidth, keepaspectratio]{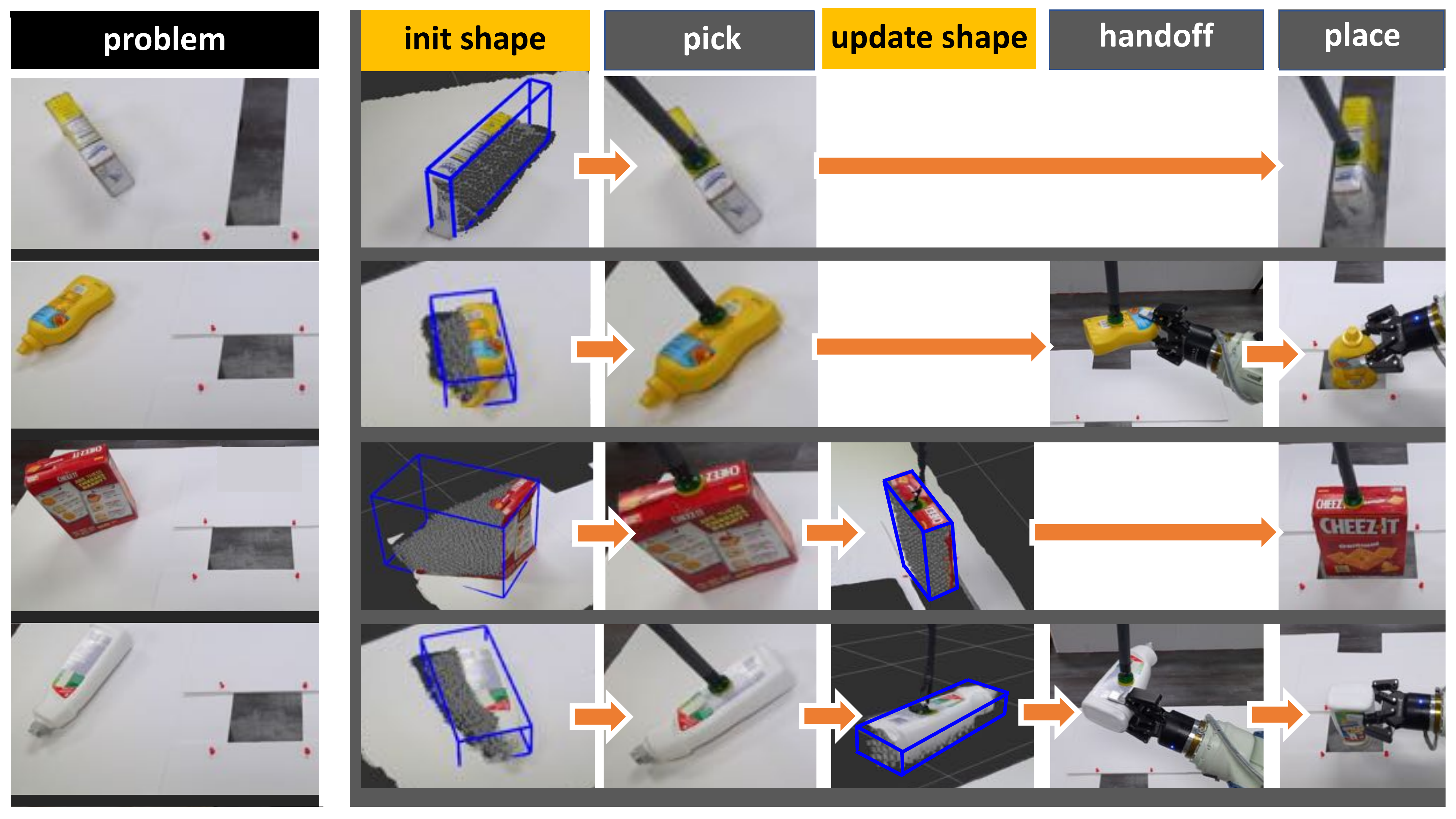}
\vspace{-0.1in}
\caption{Qualitative results indicating different solution modes of the proposed pipeline.}
\vspace{-0.2in}
\label{fig:qualitative}
\end{figure}

\begin{table*}[h]
\vspace{-0.1in}
\caption{}\vspace{-0.1in}
\centering
\begin{tabular}{lc|cc||cc||cc}
\hline
Object & \#Experiments & \multicolumn{2}{c}{Baseline} & \multicolumn{2}{c}{ (+) Handoff} & \multicolumn{2}{c}{ Proposed pipeline}\\
\hline
& & S (\%)	& S + MS (\%) &	S (\%) & S + MS (\%) & S (\%) & S + MS (\%)\\
\hline
001\_chips\_can & 20 & 15.00 & 15.00 & 35.00 & 40.00 & 90.00 & 90.00\\
003\_cracker\_box & 30 & 30.00 & 33.33 & 46.66 & 56.66 & 90.00 & 93.33\\
004\_sugar\_box & 30 & 23.33 & 23.33 & 53.33 & 60.00 & 93.33 & 96.66\\
006\_mustard\_bottle & 20 & 0.00 & 0.00 & 45.00 & 55.0 & 100.00 & 100.00\\
002\_bleach\_cleanser & 20 & 0.00 & 0.00 & 50.00 & 60.00 & 100.00 & 100.00\\
\hline
Overall & 120 & 15.83 & 16.66 & 46.66 & 55.00 & 94.16 & 95.82\\
\hline
    \multicolumn{8}{C{0.85\textwidth}}{\vspace{0.02in}Evaluating the task success rate of the proposed manipulation pipeline against a baseline. Overall 240 manipulation trials were executed, where the results corresponding to {\it Baseline} and {\it Baseline + handoff} are derived from the first set and the results for the proposed pipeline are derived from the second set. {\it S} indicates successful insertion in the constrained space, and {\it MS} stands for marginal success, where the object made contact with the boundary of the constrained space but the task still succeeded.}
\end{tabular}
\label{table:task_succ}
\end{table*}

\begin{table*}[h]
\vspace{-0.1in}
\caption{}\vspace{-0.1in}
\centering
\begin{tabular}{lccc||ccccc}
\hline
\multicolumn{4}{c}{Baseline + Handoff} & \multicolumn{5}{c}{Proposed pipeline}\\
\hline
& sense-place & sense-hoff-place & overall & place & sense-place & hoff-place & sense-hoff-place & overall\\
\hline
\#instances	& 20.0 & 46.0 & 66.0 & 18.0 & 22.0 & 51.0 & 24.0 & 115.0\\
tp time (s) & 4.29 $\pm$ 3.59 & 5.87 $\pm$ 2.88 & 5.39 $\pm$ 3.20 & 1.10 $\pm$ 0.47 & 6.69 $\pm$ 4.15 & 5.41 $\pm$ 3.14 & 13.50 $\pm$ 8.69 & 6.67 $\pm$ 6.22\\
move time (s)	& 9.92 $\pm$ 1.04 &	19.91 $\pm$ 1.87 & 16.88 $\pm$ 4.88 & 6.13 $\pm$ 2.76 & 7.24 $\pm$ 1.24 & 18.12 $\pm$ 2.02 & 18.22 $\pm$ 1.67 & 14.18 $\pm$ 5.79\\
sense actions & 4.0 $\pm$ 0.0 & 4.0 $\pm$ 0.0 & 4.0 $\pm$ 0.0 & 0.0 $\pm$ 0.0 & 1.36 $\pm$ 0.56 & 0.0 $\pm$ 0.0 & 1.41 $\pm$ 0.57 & 0.59 $\pm$ 0.81\\
\hline
\multicolumn{9}{C{0.95\textwidth}}{\vspace{0.02in} Comparing the quality and computation time for the solutions found with the baseline and the proposed approach. The data is presented only for successful executions within each category.}
\vspace{-0.1in}
\end{tabular}
\label{table:quality}
\end{table*}

\begin{figure}[h]
\vspace{-0.1in}
\centering
\includegraphics[width=0.9\linewidth, keepaspectratio]{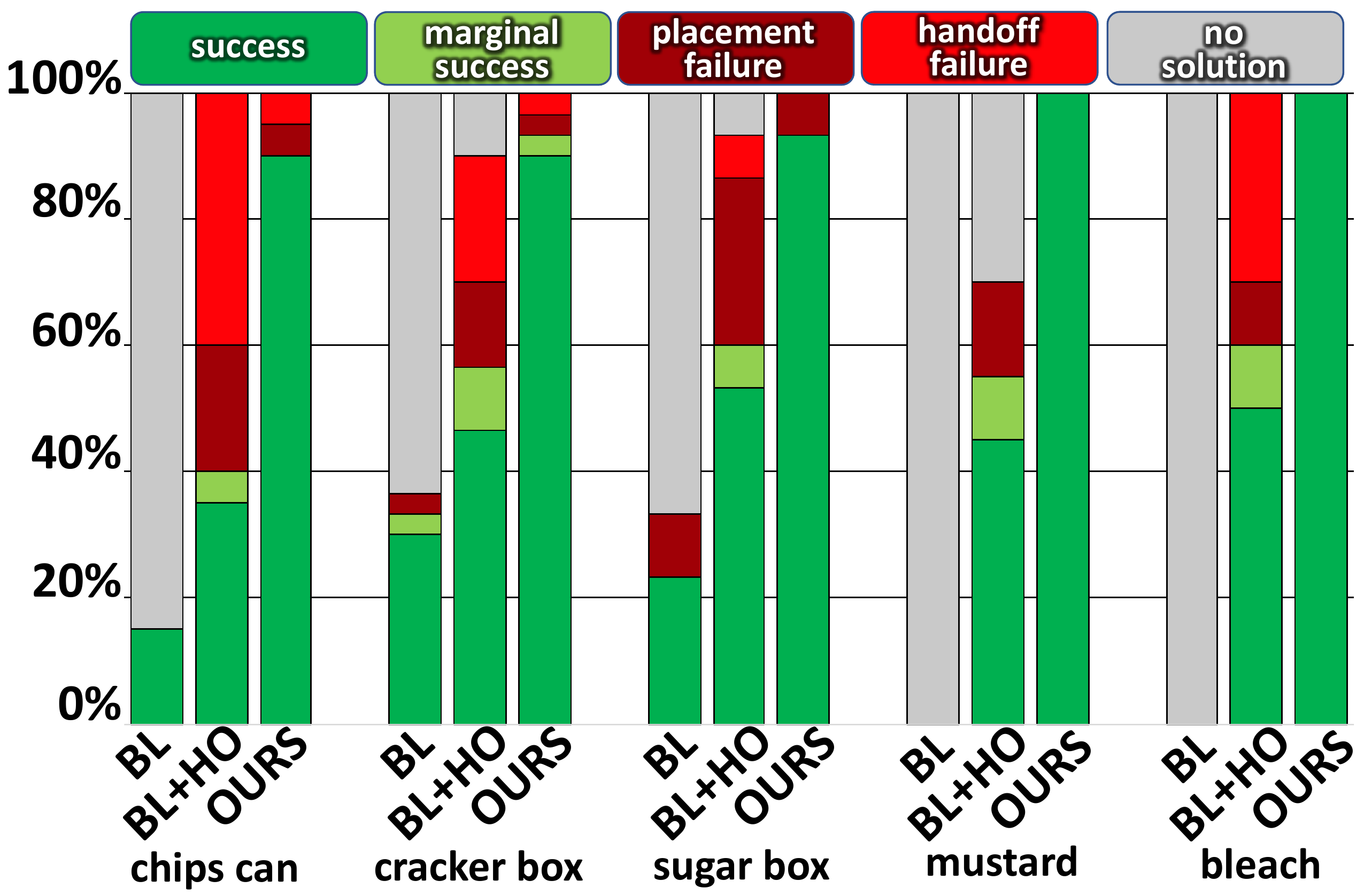}
\vspace{-0.1in}
\caption{Split of outcomes of experiments within success and various failure cases for each category.}
\vspace{-0.05in}
\label{fig:failures}
\end{figure}

$240$ trials are performed with combinations of {\it object sets}, {\it initial configurations} and {\it placement constraints}. \changed{Out of these, 120 experiments use the {\it Baseline} pipeline shown in Fig.~\ref{fig:exp_setup} and the remaining 120 use the proposed pipeline.  The results for {\it Baseline} (BL) and {\it Baseline + Handoff} (HO) are derived from the same set of physical experiments. Fig.~\ref{fig:failures} shows the outcome of the experiments.} The failures include {\it Placement failures} where the final act of placement fails to insert the object, {\it Handoff failures} where executing the transfer of object between the arms fails, and {\it No Solution} cases when planning fails and nothing is executed.

{\bf Baseline (BL): } The baseline corresponds to the shape reconstruction pipeline but \textit{without the option for handoffs}. Once picked with a task-agnostic grasp, the object is moved in front of the sensor at a predefined pose, and RGB-D images are captured from 4 different viewpoints by rotating the object along the global \textit{Z-axis} by an angle of $\pi/2$. Views are merged to obtain the object's reconstruction. Manipulation planning is then invoked to find a pick-and-placement (no handoff) solution with the left arm if it exists. The baseline achieves a very low success rate (Table.~\ref{table:task_succ}) and the most dominant failure mode is {\it No Solution} (Fig.~\ref{fig:failures}) since the initially chosen grasp might not allow task completion. \changed{This implies that given the selected grasp, a reachable, collision-free placement configuration cannot be found for the arm}.

{\bf Baseline + Handoff (HO)}: An improvement over BL, this allows the manipulator an additional option of transferring the object to the fingered gripper which can then be used to reorient and place it in the constrained space. The overall success rate increases significantly when additional handoff actions are available. Nonetheless, the handoff by itself can be seen as a constrained placement problem, and as this approach commits to a pick for object reconstruction without manipulation planning, it could still lead to {\it No solution} cases specially for relatively smaller sized objects such as for the {\it Mustard bottle} (Fig.~\ref{fig:failures}). The grasps with the fingered gripper are computed assuming that the reconstructed geometry is indeed the complete model of the object. However, views across a single rotation are not sufficient to complete the object shape. \changed{Unlike the proposed approach, the baseline does not consider the unseen part of the object as a collision geometry.} This causes grasps to collide with the unmodeled parts of the object during execution ({\it Handoff failures}). \changed{The baseline approach performs re-sensing after it picks the object. The re-sensing action prevents any inconsistency due to in-hand motion of the object during the pick. Nevertheless, any in-hand motion that occurs after the reconstruction does not get accounted for and can result in {\it Placement failures}. Placement can also fail if the reconstructed geometry is an under-approximation of the true object geometry.}

{\bf Proposed Pipeline:} The proposed pipeline discovers four classes of solutions (Fig~\ref{fig:qualitative}) that compose a sequence of \textit{picks, updates, handoffs and placements}. The key benefit is that it chooses the mode of operation based on the problem at hand, and tries to \textit{(a)} perform the minimum number of sensing actions \textit{(b)} with a minimum number of manipulation actions \textit{(c)} in a robust fashion that accounts for non-prehensile errors \textit{(d)} while guaranteeing safe execution and successful task completion. The results reflect that it achieves all of the above by leveraging the \textit{object representation}, \textit{integrated perception and planning} in the pipeline, and \textit{closed loop execution} to achieve a success rate of {\bf 95.82\%}.

The proposed pipeline eliminates the cases of {\it No Solution} by performing manipulation planning with a \textit{large, diverse, and robust} set of grasps. It ensures successful execution of the task by conservative modeling of the unseen parts of the object to avoid collision and by tracking the shape representation to account for any in-hand motion of the object and adjusting the computed plan. The failure cases for this approach are due to failures in tracking. If the within-hand motion is too drastic, motion plans might not be found for local adjustments to the initially computed solution.

As indicated in Fig.~\ref{fig:qualitative} and Table.~\ref{table:quality}, the proposed solution can find one of the four solution modes with varying solution quality. The advantage in terms of efficiency comes from the fact that the proposed solution requires additional sensing in only 38\% of the runs and the mean number of sensing actions is 1.36 as opposed to the 4 additional sensing actions in every run for the baseline approach. Additionally, the object representation allows task planning with multiple grasping options even before picking thereby increasing the number of single-shot pick-and-place solutions with less motion time. The overall execution time reduces significantly due to the combination of these factors.

\noindent{\bf Demonstrations and Publicly-shared Data: }
On top of the benchmark, additional demonstrations show the capability of the proposed system. The first demonstration is performed over mugs, some with and some without handles, with the handles being occluded in the first viewpoint. Such a case imposes ambiguity for shape completion approaches, but is solved with the proposed pipeline as demonstrated in the accompanying video. The second demonstration presents the task of flipping objects and placing them on the table. Without models, object placement tasks can either be specified relative to constraints in the environment or relative to the initial pose. Following data items corresponding to all the manipulation runs for the proposed solution are made publicly available at {\small \color{red}\url{https://robotics.cs.rutgers.edu/task-driven-perception/}}.
1) Task specification: Initial RGB-D data, object segment, placement region.
2) RGB-D data at 20Hz for the executed trajectory.
\begin{wrapfigure}{r}{0.25\textwidth}
	\vspace{-0.1in}
	\centering
	\includegraphics[width=0.25\textwidth, keepaspectratio]{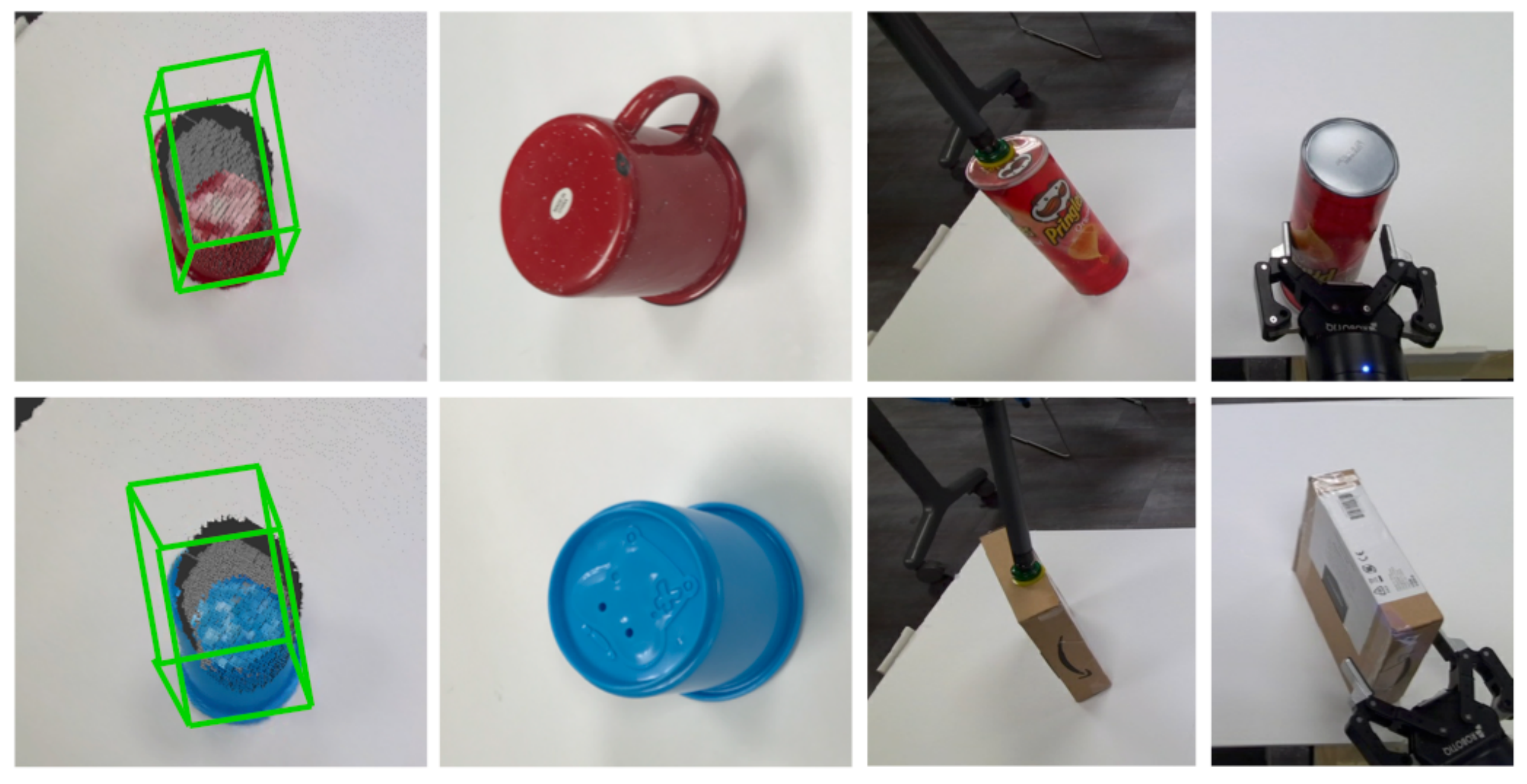}
	\vspace{-0.25in}
	\caption{Demonstrations of the proposed pipeline's operation (left) in the presence of shape ambiguity (right) on the object flipping task.}
	\vspace{-0.1in}
	\label{fig:demo}
\end{wrapfigure}
3) Robot arm transformations and grasping status for both grippers.
4) Relative pose estimates returned by the tracking module for every frame. 
The data can be used as a manipulation benchmark or to study tracking shapes and poses of objects in-hand during manipulation.

\section{Limitations and Future Work}
\label{sec:conclusion}
The current work paves the way for the paradigm of task-driven
perception and manipulation for solving pick-and-constrained-placement
tasks. Not assuming a category-level shape prior or known geometric
models and operating directly over the sensor data makes this
manipulation pipeline safe to execute and scalable. The results show
performance benefits from the design principles adopted in the
pipeline and the representation proposed in the current work.

There are some limitations to the current work that can be addressed
in future research. The pick/grasp computation is not the focus
here. General grasping strategies on such shape representations can
prove useful. \changed{Additionally, the end-effectors utilized in the
system restrict the choice of objects that can be evaluated due to
limitations based on object's weight, size or material
properties. Similar restrictions are due to the depth sensor used in
this study as it is not suited for reflective and transparent
objects.} Segmentation in the presence of clutter is
challenging \changed{despite the recent progress in depth and color
based segmentation \cite{xie2019best, danielczuk2019segmenting} of
unknown objects. Future work could focus on dealing with segmentation
noise and occlusions due to clutter.} Finally, it is often not
possible or safe to insert the object completely in a narrow opening,
and in such cases it can be dropped from some height. This process is
significantly affected by the object's mass distribution, which also
needs to be modeled.



\bibliographystyle{IEEEtran}
\bibliography{root}

\end{document}